\documentclass[10pt,twocolumn,letterpaper]{article}

\usepackage[accsupp]{axessibility}  

\usepackage{iccv}
\usepackage{times}
\usepackage{epsfig}
\usepackage{graphicx}
\usepackage{amsmath}
\usepackage{amssymb}

\usepackage{booktabs}
\usepackage{multirow}

\usepackage[pagebackref=true,breaklinks=true,letterpaper=true,colorlinks,bookmarks=false]{hyperref}

\usepackage[capitalize]{cleveref}
\crefname{section}{Sec.}{Secs.}
\Crefname{section}{Section}{Sections}
\Crefname{table}{Table}{Tables}
\crefname{table}{Tab.}{Tabs.}
\usepackage{color}
\definecolor{purple}{rgb}{0.6, 0.2, 0.8}
\definecolor{cadmiumgreen}{rgb}{0.0, 0.42, 0.24}
\definecolor{uablue}{rgb}{0.0, 0.2, 0.67}

\iccvfinalcopy 

\def\iccvPaperID{9315} 

\ificcvfinal\fi

\begin{document}

\title{LAW-Diffusion: Complex Scene Generation by Diffusion with Layouts}

\author{
Binbin Yang{$^{1}$} \quad Yi Luo{$^{1}$} \quad Ziliang Chen{$^{2}$} \quad Guangrun Wang{$^{3}$} \quad
Xiaodan Liang{$^{1}$} \quad Liang lin{$^{1}$}\thanks{Corresponding author.} \\
{\normalsize{$^{1}$}Sun Yat-Sen University \quad {$^{2}$}Jinan University \quad {$^{3}$}University of Oxford}\\
\small{\tt{yangbb3@mail2.sysu.edu.cn, luoy97@mail3.sysu.edu.cn, c.ziliang@yahoo.com,}}\\ \small{\tt{\{wanggrun, xdliang328\}@gmail.com, linliang@ieee.org}}
}

\maketitle
\ificcvfinal\thispagestyle{empty}\fi
\begin{abstract}
Thanks to the rapid development of diffusion models, unprecedented progress has been witnessed in image synthesis. Prior works mostly rely on pre-trained linguistic models, but a text is often too abstract to properly specify all the spatial properties of an image, e.g., the layout configuration of a scene, leading to the sub-optimal results of complex scene generation.
In this paper, we achieve accurate complex scene generation by proposing a semantically controllable Layout-AWare diffusion model, termed LAW-Diffusion. Distinct from the previous Layout-to-Image generation (L2I) methods that only explore category-aware relationships, LAW-Diffusion introduces a spatial dependency parser to encode the location-aware semantic coherence across objects as a layout embedding and produces a scene with perceptually harmonious object styles and contextual relations. To be specific, we delicately instantiate each object's regional semantics as an object region map and leverage a location-aware cross-object attention module to capture the spatial dependencies among those disentangled representations. We further propose an adaptive guidance schedule for our layout guidance to mitigate the trade-off between the regional semantic alignment and the texture fidelity of generated objects. Moreover, LAW-Diffusion allows for instance reconfiguration while maintaining the other regions in a synthesized image by introducing a layout-aware latent grafting mechanism to recompose its local regional semantics. To better verify the plausibility of generated scenes, we propose a new evaluation metric for the L2I task, dubbed Scene Relation Score~(SRS) to measure how the images preserve the rational and harmonious relations among contextual objects. Comprehensive experiments on COCO-Stuff and Visual-Genome demonstrate that our LAW-Diffusion yields the state-of-the-art generative performance, especially with coherent object relations.

\end{abstract}

\begin{figure}[ht!]
\begin{center}
\includegraphics[width=1.0\linewidth]{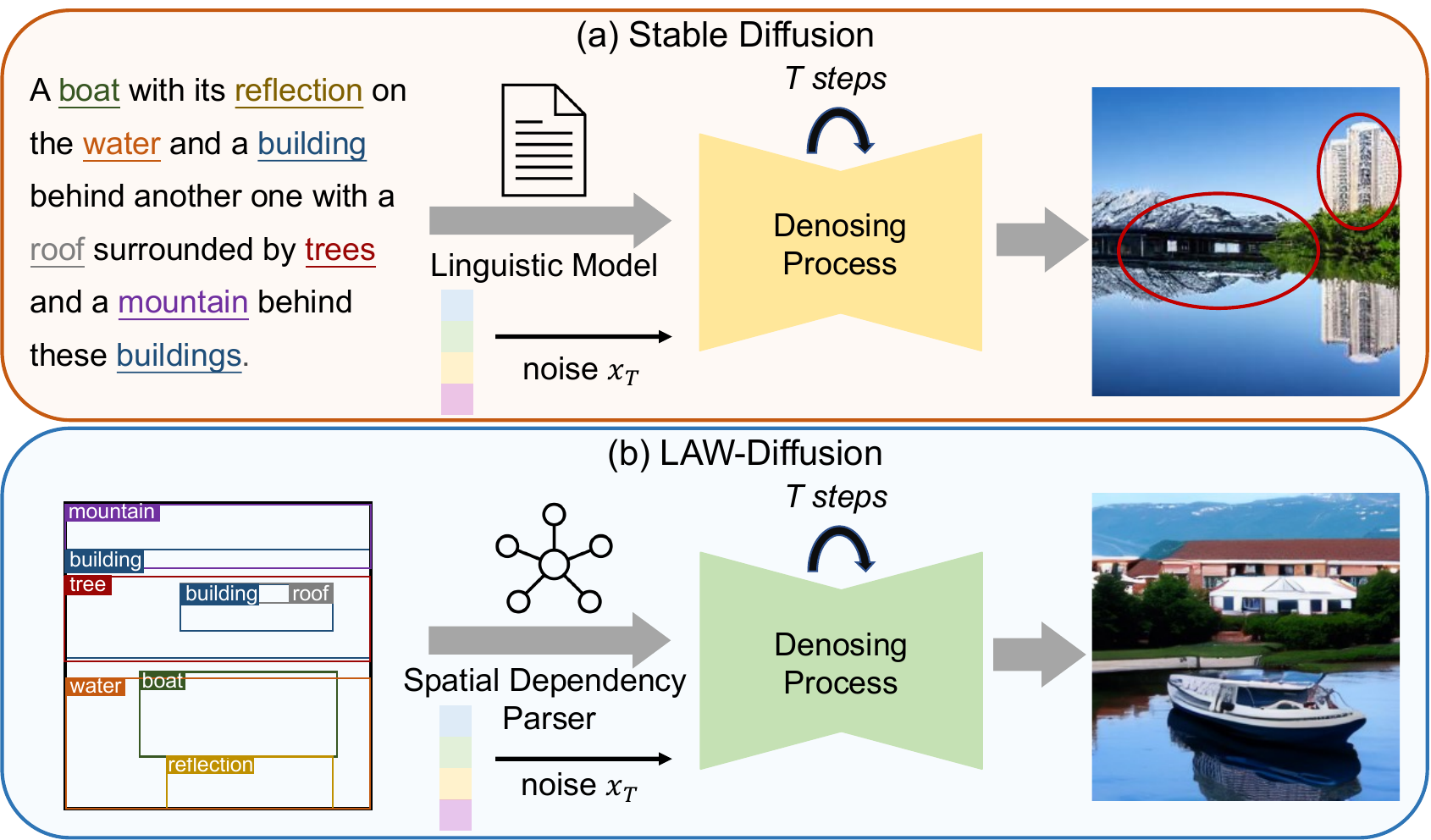}
\end{center}
\vspace{-5pt}
\caption{Illustration of complex scene generation by Stable Diffusion~\cite{rombach2022high} (text-to-image model) and our LAW-Diffusion (layout-to-image model). Stable Diffusion relies on linguistic model and generates an unsatisfactory scene: the \textcolor{cadmiumgreen}{boat} on the water is missed and the generated \textcolor{uablue}{building} and \textcolor{purple}{mountain} are placed with undesired spatial relation according to the input description. By contrast, LAW-Diffusion introduces a spatial dependency parser to encode the spatial semantic coherence and produces the scene image with consistent contextual relations adhere to the layout configuration.
}
\label{fig:fig_t2i_l2i}
\vspace{-5pt}
\end{figure}

\section{Introduction}
\label{sec:intro}
Recently, astounding advances have been achieved in generative modeling due to the emergence of diffusion models~\cite{sohl2015deep, ho2020denoising, rombach2022high, yang2022reco, avrahami2022blended, fan2022frido}. Despite the stunning generative performance in simple cases, \emph{e.g.}, single object synthesis, how to generate a complex scene composed of multiple visual concepts with their diverse relationships remains a challenging problem. A straightforward solution is to translate the scene into a text description and then resort to the state-of-the-art text-to-image~(T2I) generative models~\cite{rombach2022high, fan2022frido, gafni2022make, saharia2022photorealistic, ramesh2022hierarchical}. However, text-to-image diffusion models, \emph{e.g.}, Stable Diffusion and its variants~\cite{rombach2022high, fan2022frido, gafni2022make, saharia2022photorealistic, ramesh2022hierarchical} fall short when it comes to the spatial composition of multiple objects in a scene. An underlying reason is that properly specifying all the spatial properties
in an abstractive sentence is laborious and less accurate, usually resulting in unsatisfactory generated results. In addition, the linguistic model used in T2I model is incapable of accurately capturing the objects' spatial relations whereas only providing a coarse-grained linguistic understanding from the text description. An example is shown in \cref{fig:fig_t2i_l2i}, in which we extract a sentence description from a scene layout configuration and compare the generated results of Stable Diffusion~\cite{rombach2022high} and our model.
From the result generated by Stable Diffusion in \cref{fig:fig_t2i_l2i}(a), we can observe that the spatial properties are not well preserved (\emph{e.g.}, the generated mountain is besides the building while it should be behind the building) and some desired objects are missed~(\emph{e.g.}, the boat and its reflection). By contrast, our method generates the scene image by directly parsing the spatial dependency in the layout configuration.

Layout-to-image generation~(L2I) is a very important task of controllable image synthesis, which takes a configuration of visual concepts (\emph{i.e.}, objects' bounding boxes with their class labels in a certain spatial layout) as the input. The scene layout precisely specifies each object's size, location and its association to other objects. The key challenge for L2I lies in encoding the spatial dependencies among co-existing objects at each position, \emph{i.e.}, the location-aware semantic composition, which is vital to eliminate the artifacts of spurious edges between spatial adjacent or overlapped objects~\cite{he2021context}.
Existing studies on L2I are usually developed based on the generative adversarial networks~(GAN)~\cite{goodfellow2020generative, sun2021learning, he2021context, sylvain2021object, zhao2019image}. These methods render the realism of image contents with instance-specific style noises and discriminators, and thus suffer from the lack of overall 
harmony and style consistency among things and stuffs in the generated scene. They have made a few attempts to capture the class-aware relationships in the generator by adopting LSTM~\cite{zhao2019image} or attention mechanism~\cite{he2021context}.
Another type of approaches is based on transformer~\cite{jahn2021high, yang2022modeling}, which reformulates the scene generation task as a sequence prediction problem by converting the input layout and target image into a list of object tokens and patch tokens. The transformer~\cite{vaswani2017attention} is then employed to sequentially predict the image patches, which actually capture the sequential dependencies rather than scene coherence. Recently, generic T2I diffusion models, \emph{e.g.}, LDM~\cite{rombach2022high} and Frido~\cite{fan2022frido} have been demonstrated that they can be extended to L2I by tokenizing the layout into a sentence-like sequence of object tokens and encoding them by linguistic model, following their standard T2I paradigm. Such brute-force solutions share some shortcomings inherent to the T2I diffusion models, \emph{e.g.}, the aforementioned object leakage and unawareness of spatial dependencies in \ref{fig:fig_t2i_l2i}(a).
But in fact, prior methods mainly exploit the location-insensitive relationships while overlooking the fine-grained location-aware cross-object associations.

To address the above issues, we present a novel diffusion model-based framework for L2I, termed \emph{LAW-Diffusion}, for synthesizing complex scene images with mutually harmonious object relations. Unlike the traditional L2I methods treating each object separately, our LAW-Diffusion learns a layout embedding with rich regional composition semantics in a delicate manner for better exploring the holistic spatial information of objects. Concretely, we first instantiate each object's regional semantics as an object region map that encodes the class semantic information in its bounding box. Then, we split those region maps into fragments and propose a location-aware cross-object attention module to perform per-fragment multi-head attention with a learnable aggregation token to exploit the location-aware composition semantics. By regrouping those aggregated fragments according to their original spatial locations, we obtain a layout embedding encapsulating both class-aware and location-aware dependencies. In this way, when synthesizing a local fragment of image, such composed semantics faithfully specify whether objects are possibly overlapped at the certain location. Inspired by the effectiveness of text-to-image diffusion models~\cite{ramesh2022hierarchical, saharia2022photorealistic, nichol2021glide}, we employ the form of classifier-free guidance~\cite{ho2022classifier} to amplify the regional control from our layout embedding. To avoid losing objects' texture details when leveraging a large guidance scale, we further propose an adaptive guidance schedule for the sampling stage of LAW-Diffusion to maintain both layout semantic alignment and object's texture fidelity by gradually annealing the guidance magnitude. Furthermore, LAW-Diffusion allows for instance reconfiguration, \emph{e.g.}, adding/removing/restyling an instance in a generated scene via layout-aware latent grafting. Specifically, we spatially graft an exclusive region outside a bounding box from the diffusion latent of the already generated image onto the target latent guided by a new layout at the same noise level. By alternately recomposing the local regional semantics
and denosing these grafted latents, LAW-Diffusion can reconfigure an instance in a synthesized scene image while keeping the other objects unchanged.

The existing evaluation metrics for the L2I task basically focus on measuring the fidelity of generated objects while ignoring the coherence among objects' relations in the scene context. Thus, we propose a new evaluation metric called Scene Relation Score~(SRS) to measure whether the generated scenes preserve the rational and harmonious relations among contextual objects, which would facilitate the development of L2I research. We conduct both quantitative and qualitative experiments on Visual Genome~\cite{krishna2017visual} and COCO-Stuff~\cite{caesar2018coco}, and the experimental results demonstrate that our LAW-Diffusion outperforms other L2I methods and achieves the new state-of-the-art generative performance, particularly in preserving reasonable and coherent object relations.

\begin{figure*}[ht!]
\begin{center}
\includegraphics[width=0.95\linewidth]{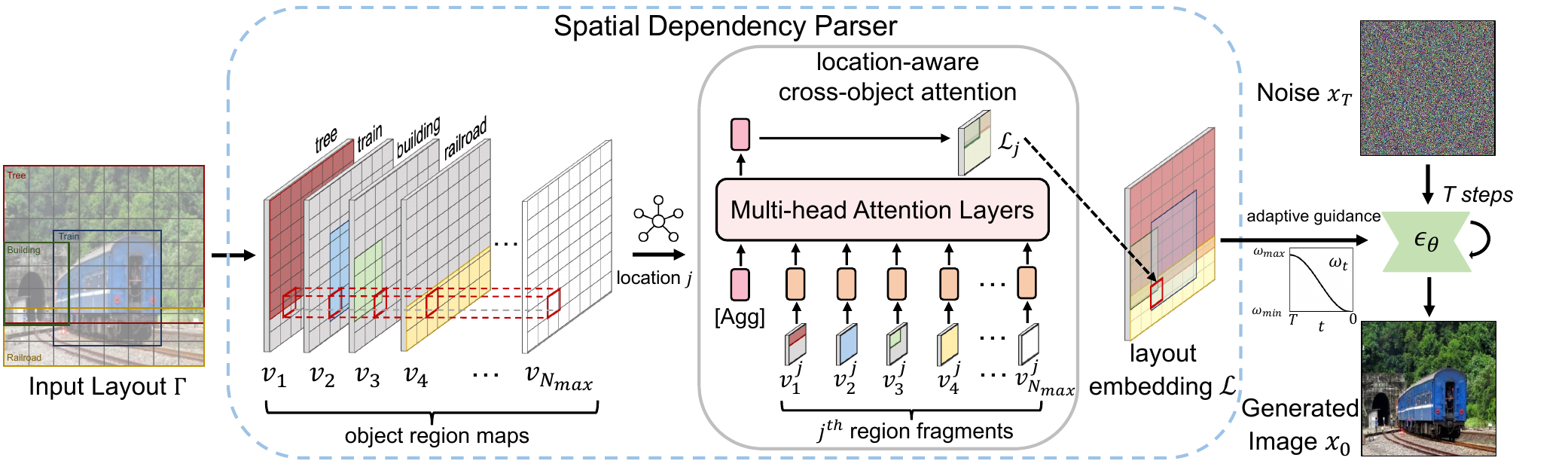}
\end{center}
\caption{An overview of LAW-Diffusion. Given an input layout $\Gamma$, each object's region map $v_i$ is generated as its regional semantics by filling its class embedding into the region specified by its bounding box. The object region maps are split into patches of region fragments. For the region fragments at the location $j$, the location-aware cross-object attention module is used to aggregate them as $\mathcal{L}_j$ via multi-head attention. In this way, $\mathcal{L}_j$ encodes the spatial dependencies among objects at this location. Furthermore, the layout embedding $\mathcal{L}$ is obtained by collecting all aggregated fragments and used to control the generation of LAW-Diffusion with an adaptive guidance schedule: the guidance magnitude $\omega_t$ gradually anneals from $\omega_{\max}$ to $\omega_{\min}$ during denoising process. Best viewed in color.}
\label{fig:overview}
\end{figure*}

\section{Related Work}
\label{sec:relate}
\noindent \textbf{Diffusion Models} \;Diffusion models~\cite{sohl2015deep, ho2020denoising, luo2021diffusion, lugmayr2022repaint, saharia2022image} recently emerges as powerful image generators due to their impressive generative performance.
By training a noise estimator, the generative process of diffusion model is formulated as iteratively denoising from an image-level noise~\cite{ho2020denoising, dhariwal2021diffusion}. With the introduction the techniques of classifier guidance~\cite{dhariwal2021diffusion} and classifier-free guidance~~\cite{ho2022classifier}, diffusion models are enabled to incorporate different types of conditional information during the sampling stage. Most recent progresses~\cite{avrahami2022blended,yang2022reco, fan2022frido, ruiz2022dreambooth, gal2022image, gafni2022make, saharia2022photorealistic, ramesh2022hierarchical} are made in the field of text-to-image (T2I) generation because the prevalence of Stable Diffusion~\cite{rombach2022high}. However, those T2I diffusion models always fall short when it comes to the complex spatial semantic composition of multiple objects in a scene. In this paper, we manage to present a layout-aware diffusion model for complex scene image generation, by mining the spatial dependencies among co-existing objects in the scene layout.

\noindent \textbf{Layout-to-Image Generation} \;Image generation from a layout configuration~(L2I) is a specific task of conditional image generation, whose input is a set of bounding boxes and class labels of the objects in a scene. It liberates people from racking their brains to formulate an accurate but complicated language description of a complex scene and rather provides a more flexible human-computer interface for scene generation. Layout2Im~\cite{zhao2019image} generated objects' features from noises and class labels, and fused them by LSTM~\cite{hochreiter1997long}. LostGAN~\cite{sun2021learning} further introduced mask prediction as an intermediate process and proposed an instance-specific normalization to transform the object features. OC-GAN~\cite{sylvain2021object}, Context-L2I~\cite{he2021context} and LAMA~\cite{li2021image} followed their training schemes and further improved objects' representations and the quality of mask generation. Transformer based methods~\cite{jahn2021high, yang2022modeling} converted the layout and image into object tokens and patch tokens, which reformulating L2I as a sequence prediction task. Recently, T2I diffusion models~\cite{rombach2022high, fan2022frido} are extended to L2I through encoding the list of object tokens by linguistic model and then regarding it as a T2I task. However, prior approaches merely mine the category-aware retionships while overlooking the location-aware cross-object associations. In this work, we present LAW-Diffusion by explicitly encoding the location-aware semantic compositions for the visual concepts in the scene.

\section{LAW-Diffusion}
\label{sec:method}
\subsection{Preliminaries}
\noindent \textbf{Diffusion Models} \; Diffusion model is a type of likelihood-based generative models, consisting of a forward diffusion process and a backward denoising process. Formally, given an image sample $x_0 \sim q(x_0)$, the forward process is defined as a Markov chain with Gaussian transitions:
\begin{equation}
    q(x_t | x_{t-1}) = \mathcal{N}(x_t; \sqrt{\alpha_t}x_{t-1}, (1-\alpha_t) \mathbf{I}),
\end{equation} 
where $\{\alpha_t \in (0,1)\}_{t=1}^T$ is a deceasing sequence of the noise magnitudes in each step. From the property of Gaussian noise and Markov chain, we can directly derive the transition from $x_0$ to any latent variable $x_t$:
\begin{equation}
    q(x_t | x_0) = \mathcal{N}(x_t; \sqrt{\bar{\alpha}_t}x_0, (1-\bar{\alpha}_t)\mathbf{I}),
\end{equation}
where $\bar{\alpha}_t = \prod_{s=1}^T\alpha_s$. By re-parameterization, $x_t$ can be written as the weighted sum of $x_0$ and a noise $\epsilon \sim \mathcal{N}(\mathbf{0}, \mathbf{I})$:
\begin{equation}
\label{eq:reparam}
    x_t = \sqrt{\bar{\alpha}_t}x_0 + \sqrt{1-\bar{\alpha}_t}\epsilon.
\end{equation}
A simple conclusion is that if the length of the Markov chain $T$ is large enough, $\bar{\alpha}_T \approx 0$ and $x_T$ will approximately follow a standard Gaussian distribution $\mathcal{N}(\mathbf{0}, \mathbf{I})$. 

The generative process of diffusion model is defined as iteratively denoising from the Gaussian prior, \emph{i.e.}, $x_T \sim \mathcal{N}(\mathbf{0}, \mathbf{I})$. Due to the intractability of the reverse transition $q(x_{t-1}|x_t)$, another Markov process parameterized by $\theta$, \emph{i.e.}, $p_{\theta}(x_{t-1}|x_t)$ is learned to serve as its approximation and generate the denoised results $\{x_T, x_{T-1}, ..., x_0\}$:
\begin{equation}
    p_{\theta}(x_{t-1}|x_t) = \mathcal{N}(x_{t-1}; \mu_{\theta}(x_t, t), \Sigma_{\theta}(x_t, t)).
\end{equation}
Denoising diffusion probabilistic models (DDPM)~\cite{ho2020denoising} reveal that $\mu_\theta(x_t, t)$ derives from a noise estimator $\epsilon_{\theta}(x_t, t)$:
\begin{equation}
    \mu_{\theta}(x_t, t) = \frac{1}{\sqrt{\alpha_t}}\left(x_t - \frac{1-\alpha_t}{1-\bar{\alpha}_t}\epsilon_{\theta}(x_t, t)\right).
\end{equation}
By optimizing the re-weighted variational lower-bound (VLB) on $\log p_{\theta}(x_0)$~\cite{ho2020denoising}, the noise estimator $\epsilon_{\theta}(x_t, t)$ is trained to predict the noise $\epsilon$ in \cref{eq:reparam} and enables diffusion models to produce image samples:
\begin{equation}
\label{eq:vlb}
\small
    L_{\text{VLB}}(\theta) = \mathbb{E}_{t\sim[1,T], x_0 \sim q(x_0), \epsilon\sim\mathcal{N}(\mathbf{0}, \mathbf{I})}\left[\Vert\epsilon_{\theta}(x_t, t) - \epsilon\Vert^2\right].
\end{equation}

\noindent \textbf{Conditional Diffusion Models} \; Classifier-guidance~\cite{dhariwal2021diffusion} provides a way for diffusion model to achieve conditional generation by using the gradient of a separately trained classifier $p(y|x_t)$ during sampling. As a more efficient technique, classifier-free guidance~\cite{ho2022classifier, nichol2021glide} replaces the noise estimator by a combination of conditional and unconditional model, without requirement of $p(y|x_t)$:
\begin{equation}
\label{eq:vanilla_classifer_free}
    \tilde{\epsilon}_{\theta}(x_t, t|y) = \omega\epsilon_{\theta}(x_t, t|y) + (1-\omega)\epsilon_{\theta}(x_t, t|\emptyset),
\end{equation}
where $y$ is the class label or text embedding from language model~\cite{nichol2021glide}, $\omega \geq 1$ denotes the guidance scale and trivially increasing $\omega$ will amplify the effect of conditional input.

With the help of large-scale pre-trained CLIP~\cite{radford2021learning} and other language models~\cite{saharia2022photorealistic}, diffusion models produce impressive results on text-to-image generation. However, their performance of complex scene generation are always unsatisfactory because the text embeddings from the linguistic models can not accurately capture the spatial properties, \emph{e.g.}, objects' locations,  sizes and their implicit spatial associations. Distinct from text prompts, we focus on the task of generating complex scene images from the structured layout configurations~(L2I) and further propose a diffusion model-based method with flexibility and compositionality.

\begin{figure*}[ht!]
\begin{center}
\includegraphics[width=0.9\linewidth]{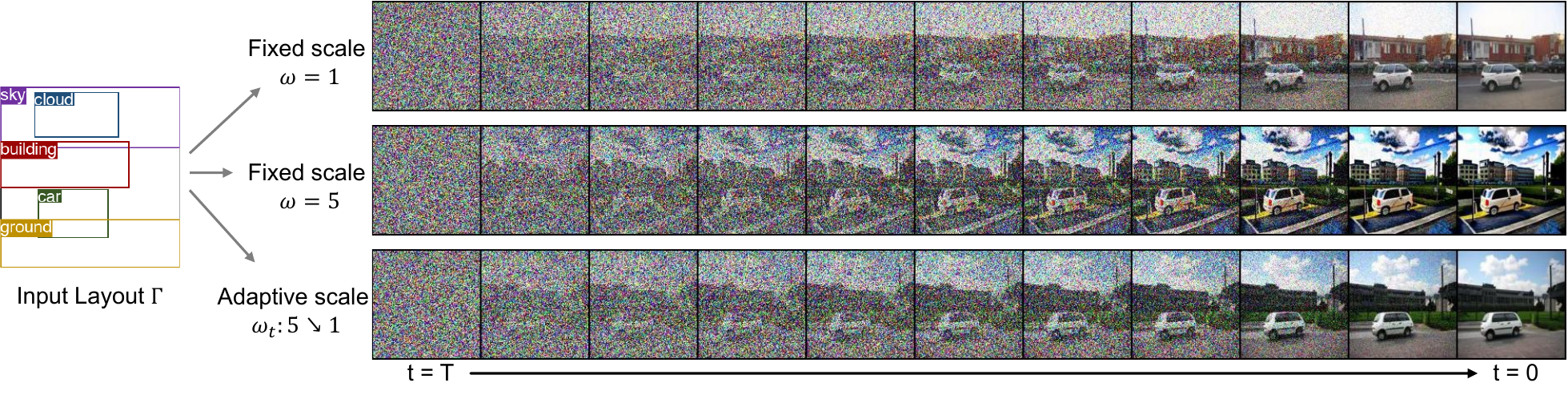}
\end{center}
  \caption{Illustration of the generation processes from the same input layout $\Gamma$ using different guidance scales. A fixed small scale $\omega=1$ for each denoising step provides insufficient semantic control, and the cloud is missed in the first row. In the second row, using a fixed large scale $\omega=5$ leads to over-saturation and distortion of object texture. In the third row, using the adaptive guidance scale $\omega_t: 5\tiny{\searrow}1$ which anneals from $\omega_T = 5$ to $\omega_1=1$ maintains both semantic alignment and photo-realism. Best viewed in color.}
\label{fig:scale_motivation}
\end{figure*}

\subsection{Layout-aware Diffusion Model}
\label{sec:LAW}
In this section, we propose a \textbf{L}ayout-\textbf{AW}are diffusion model~(LAW-Diffusion) to parse the spatial dependencies among co-existing objects and generate photo-realistic scene images with regional semantic alignment. The overview of our LAW-Diffusion is illustrated in \cref{fig:overview} and we will elaborate the details following.

\noindent \textbf{Layout-to-Image Generation} \; Complex scene image synthesis from layout configuration, also known as layout-to-image generation, is specified by synthesising an image $x \in \mathbb{R}^{H\times W\times 3}$ satisfying a layout configuration $\Gamma$ consisting of $N$ objects $\mathcal{O} = \{o_1,o_2, ...,o_{N}\}$. Each object $o_i$ is equipped with its bounding box $b_i = [r_x^i, r_y^i, h_i, w_i]$ and category $c_i$, where $(r_x^i, r_y^i)$ is the left-top coordinate and $(h_i, w_i)$ represents the object size.

\noindent \textbf{Spatial Dependency Parser} \;Unlike existing diffusion-based L2I solutions that depends on linguistic models~\cite{rombach2022high, fan2022frido}, LAW-Diffusion explores a distinctive way to explicitly harvest both location-aware and category-aware object dependencies in the compositional configurations by a spatial dependency parser. The parsing process is detailed below.

Aiming at condensing each object's spatial localization and class information, we first instantiate the regional semantics of object $o_i$ as an object region map $v_i \in \mathbb{R}^{H \times W \times d_{c}}$, which shares the same spatial resolution as image $x$ for spatial location alignment. Concretely, the rectangular region in $v_i$ specified by the bounding box $b_i$ is filled with a learnable class embedding $c_i\in \mathbb{R}^{d_c}$~(for brevity, symbol $c_i$ is reused here), while the exclusive area is filled by a learnable background embedding $c_{bg} \in \mathbb{R}^{d_c}$.
Since the number of objects $N$ varies in different layout configurations, the set of region maps $\{v_i\}_{i=1}^N$ is padded to $\{v_i\}_{i=1}^{N_{\max}}$ using a learnable null region map $v_{\emptyset}\in \mathbb{R}^{H \times W \times d_{c}}$, where $N_{\max}$ denotes the maximum number of objects.

In order to fully exploit the spatial dependencies among objects at each position, we propose a location-aware cross-object attention module to aggregate those disentangled object region maps $\{v_i\}_{i=1}^{N_{\max}}$ by their location-aware semantic composition. We split each object region map $v_i$ into $N_p$ patches of region fragments $\{v_i^{j}\}_{j=1}^{N_p}$, $v_i^j \in \mathbb{R}^{P\times P\times d_c}$ and perform multi-head self-attention~(MHSA) for the set of region fragments at the same location. Formally, for the position of the $j^{th}$ patch, we formulate $\{v_i^j\}_{i=1}^{N_{\max}}$ as an unordered set of $N_{\max}$ objects' $j^{th}$ region fragments and feed them into the stacked $L$ multi-head attention~\cite{vaswani2017attention} layers with a learnable aggregation token $v_{\text{[Agg]}}\in \mathbb{R}^{P\times P\times d_c}$: 
\begin{equation}
    z_j^0 = \text{concat}([v_{\text{[Agg]}}, v_1^{j}, v_2^j, ..., v_{N_{\max}}^{j}];
\end{equation}
\begin{equation}
\label{eq:mhsa}
    \tilde{z}_j^l = \text{MHSA}(\text{LN}(z_j^{l-1})) + z_j^{l-1}, l = 1,...,L;
\end{equation}
\begin{equation}
    z_j^l = \text{MLP}(\text{LN}(\tilde{z}_j^l)) + \tilde{z}_j^l, l = 1, ..., L;
\end{equation}
\begin{equation}
    \mathcal{L}_j = \text{LN}(z_j^L)[0],
\end{equation}
where $\mathcal{L}_j$ is the composed regional semantics for the $j^{th}$ patch. In this way, the per-fragment multi-head self attention in \cref{eq:mhsa} serves as a location-specific permutation-equivariant interaction between different objects' representations. Furthermore, by regrouping $\{\mathcal{L}_j\}_{j=1}^{N_p}$ according to their original spatial locations, we obtain the layout embedding $\mathcal{L}$ with abundant spatial dependencies among objects.

\noindent \textbf{Layout Guidance} \;To develop a diffusion model with flexible control, we train LAW-Diffusion with the classifier-free guidance~\cite{ho2022classifier, nichol2021glide} from the learned layout embedding $\mathcal{L}$, which contains regional composition semantics. Similar to \cref{eq:vanilla_classifer_free}, LAW-Diffusion learns a noise estimator $\tilde{\epsilon}_\theta(x_t, t|\mathcal{L})$ conditioned on the layout embedding $\mathcal{L}$:
\begin{equation}
\label{eq:classifer_free}
    \tilde{\epsilon}_{\theta}(x_t, t|\mathcal{L}) = \omega\epsilon_{\theta}(x_t, t|\mathcal{L}) + (1-\omega)\epsilon_{\theta}(x_t, t|\emptyset),
\end{equation}
where $\omega \geq 1$ denotes the magnitude of the layout guidance.

According to the spatial inductive bias of the image-level noise $x_T$ introduced by diffusion models, we concatenate the noised latent code $x_t$ and the layout embedding $\mathcal{L}$ to align their spatial information, \emph{i.e.}, $\text{concat}([x_t, \mathcal{L}]) \in \mathbb{R}^{H\times W\times (D+3)}$ and use it as the input of the conditional noise estimator in \cref{eq:classifer_free}:
\begin{equation}
\label{eq:epsilon_concat}
    \epsilon_{\theta}(x_t, t|\mathcal{L}) = \epsilon_{\theta}(\text{concat}([x_t, \mathcal{L}]),t),
\end{equation}
where $\epsilon_{\theta}$ is implemented by a U-Net~\cite{ronneberger2015u} and $t$ is implemented as a sinusoidal time embedding following~\cite{ho2020denoising}.

To this end, the layout embedding $\mathcal{L}$ encapsulates location-aware semantic composition of the multiple visual concepts in the scene. By absorbing the nutrition from $\mathcal{L}$ using the classifier-free guidance, LAW-Diffusion is able to generate a scene image with accurate regional semantics adhere to the input layout and coherent object relations.

\subsection{Adaptive Guidance Schedule}
As previously discussed, classifier-free guidance~\cite{nichol2021glide} provides a effective way to improve the semantic control during the sampling stage. The vanilla classifier-free guidance uses a fixed guidance scale $\omega$ in \cref{eq:classifer_free} for each denoising step $t$ and has shown its effectiveness in a variety of application scenarios~\cite{nichol2021glide, saharia2022photorealistic, ramesh2022hierarchical}. However, we empirically find in our experiment that the fixed $\omega$ will result in a dilemma of the trade-off between the layout semantic alignment and the photo-realism of generated objects. As shown in \cref{fig:scale_motivation}, a fixed small guidance scale~($\omega=1$) offers insufficient semantic control, \emph{e.g.}, the cloud is missed, while a strong guidance~($\omega=5$) leads to an over-saturated image where the cloud and car have over-smooth textures. Based on these observations, we can intuitively conclude that a large $\omega$ provides precise semantic compliance with the layout $\Gamma$ while a small $\omega$ encourages photo-realistic textures for objects. Inspired by the human's instinct of first conceiving the holistic semantics and then refining the details when drawing a picture, we propose an adaptive guidance schedule to mildly mitigate the aforementioned trade-off.

Specifically, our proposed adaptive guidance schedule is to gradually anneal the guidance magnitude $\omega_t$ during the sampling process of LAW-Diffusion: the generation starts with an initially large guidance scale $\omega_T = \omega_{\max}$ and it gradually anneals to a small magnitude $\omega_1 = \omega_{min}$ with the annealing function $\phi(t)$~($t$ is decreasing from $T$ to $1$ in the sampling stage):
\begin{equation}
    \omega_t = \omega_{\min} + \phi(t)(\omega_{\max} - \omega_{\min}).
\end{equation}
For simplicity, here we specify $\phi(t)$ as the cosine-form annealing, due to its concave property in the early denoising steps:
\vspace{-2mm}
\begin{equation}
\label{eq:cosine_adaptive}
    \omega_t = \omega_{\min} + \frac{1}{2}\left(1 + \cos(\frac{T-t}{T}\pi)\right)(\omega_{\max} - \omega_{\min}).
\end{equation}

In \cref{fig:scale_motivation}, it is evident that the adaptive guidance scale $\omega_t$ annealing from $\omega_T=5$ to $\omega_1=1$ (denoted as $\omega_t: 5\tiny{\searrow}1$) combines the benefits of the fixed guidance with $\omega=5$ and $\omega=1$, thus enabling both accurate layout semantic alignment and preservation of photo-realistic textures.

\begin{figure}[th]
\begin{center}
\includegraphics[width=1.0\linewidth]{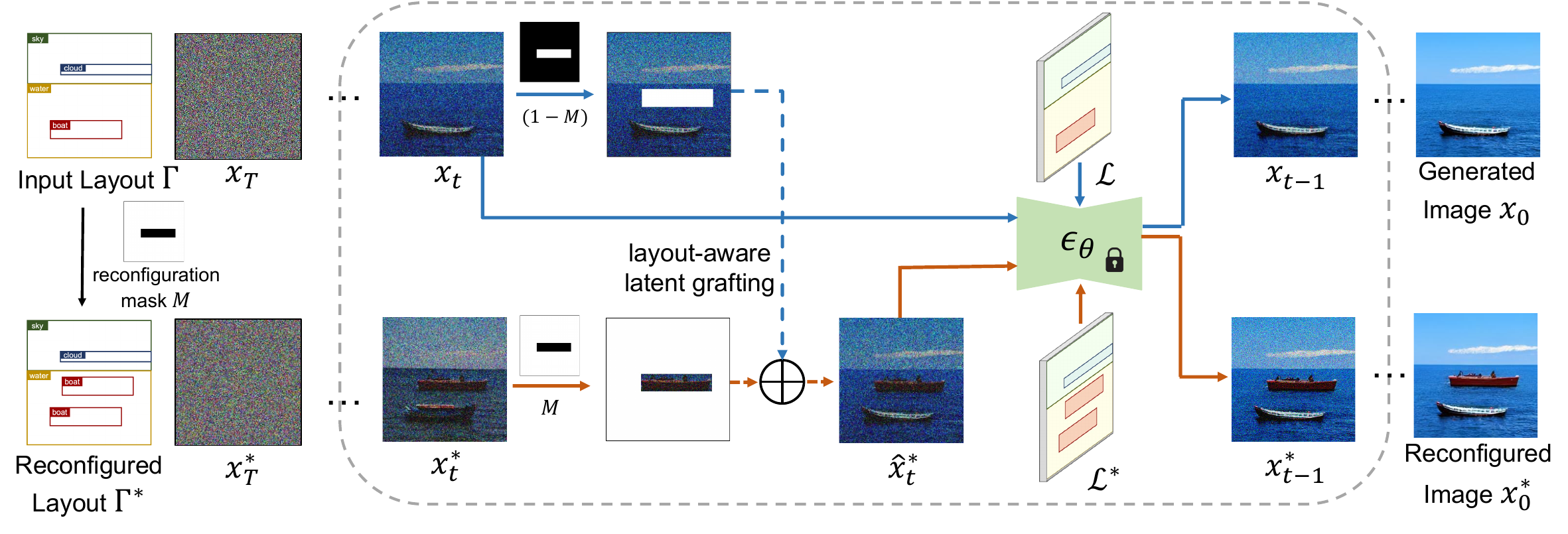}
\end{center}
\caption{Illustration of our layout-aware latent grafting mechanism for instance reconfiguration (adding an object is taken as an example). Given an image $x_0$ generated from the layout $\Gamma$, reconfigured $x_0^*$ is obtained by alternately grafting the region outside the object bounding box from $x_t$ to $x_t^*$ ($\hat{x}_t^*$ is produced), and denoising the grafted latent $\hat{x}_t^*$ to $x_{t-1}^*$ with the guidance of a reconfigured layout $\Gamma^*$. Mask $M$ indicates the region within the bounding box.}
\label{fig:graft}
\end{figure}

\subsection{Layout-aware Latent Grafting}
\label{sec:reconfigure}

To further explore the semantic controllability, we will showcase that LAW-Diffusion is capable of instance-level reconfiguration.
Although LAW-Diffusion does not explicitly model each instance's style by an individual noise like previous works~\cite{goodfellow2020generative, sun2021learning, he2021context, sylvain2021object, zhao2019image}, it allows for adding/removing/restyling an instance in the generated scene image by introducing a training-free layout-aware latent grafting mechanism. \cref{fig:graft} illustrates the process.

Formally, suppose a scene image $x_0$ has been synthesized from the layout configuration $\Gamma$ by learning its layout embedding $\mathcal{L}$, the process of instance reconfiguration can be formulated as generating an image $x^*_0$ from another configuration $\Gamma^*$ with layout embedding $\mathcal{L}^*$, where an object $o^*$ within a bounding box $b^*$ is added/removed/restyled while preserving the other objects in $x_0$. Inspired by the grafting technique used in horticulture~\cite{mudge2009history, melnyk2015plant} which connects the tissue of a plant to another plant and make them grow together, we aim to spatially graft the exclusive region outside $b^*$ from the latents $\{x_t\}_{t=1}^T$ guided by $\mathcal{L}$ onto the target latents $\{x_t^*\}_{t=1}^T$ guided by $\mathcal{L^*}$ at the same noise level. The reconfiguration process is performed by alternately grafting from $x_t$ to $x_t^*$ and denoising $\hat{x}_t^*$ to $x_{t-1}^*$:
\begin{equation}
\left\{
\begin{array}{ll}
\hat{x}_t^* = x_t^*\odot M \oplus x_t\odot (1-M),\\
x_{t-1}^* \sim p_{\theta}(x_{t-1}^*|\hat{x}_{t}^*, \mathcal{L^*}),
\end{array}
\right.
\end{equation}
where $\odot$ and $\oplus$ denotes element-wise multiplication and addition, $M$ denotes a rectangular mask indicating the region within the bounding box $b^*$, $\hat{x}_t^*$ is the grafted latent, $p_{\theta}(x_{t-1}^*|\hat{x}_{t}^*, \mathcal{L^*})$ denotes the layout-aware denoising process guided by $\mathcal{L^*}$, $x_T^*$ is initialized as a Gaussian noise distinct from $x_T$. Since $x_t^*$ is guided by holistic semantics from $L^*$ instead of only local control within $b^*$, LAW-Diffusion is able to yield a reconfigured scene with coherent relations.

\begin{figure*}[ht!]
\begin{center}
\includegraphics[width=0.96\linewidth]{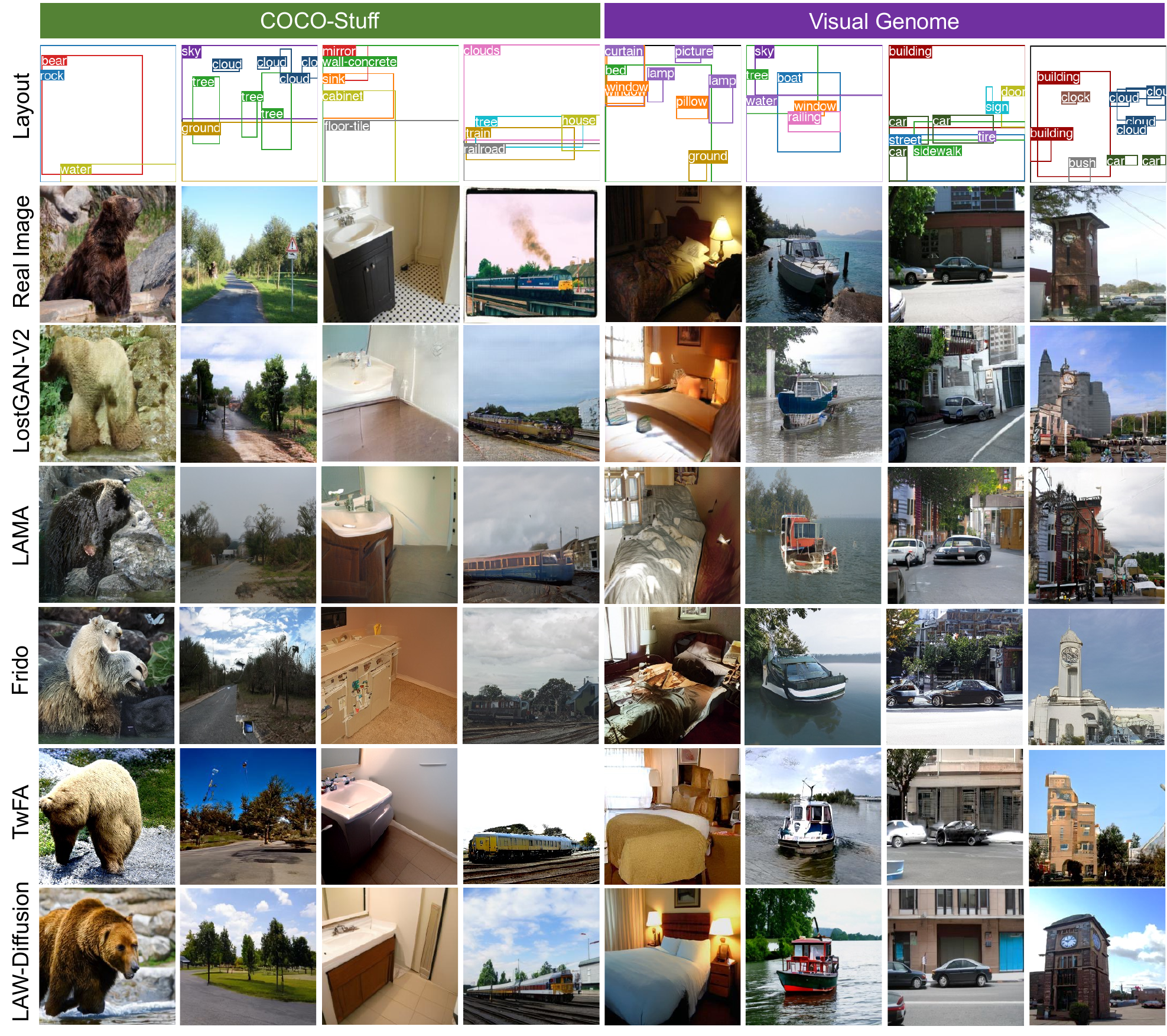}
\end{center}
\caption{Examples of the $256\times 256$ images generated by different layout-to-image methods on COCO-Stuff~\cite{caesar2018coco} and Visual Genome~\cite{krishna2017visual}. The first row shows the visualizations of layout configurations and the sampled images in the same column share a common input layout.}
\label{fig:visual_comparison}
\end{figure*}

\section{Experiments}
\subsection{Experimental Settings}
\noindent\textbf{Datasets} \;
Following existing works on layout-to-image generation, our experiments are conducted on two benchmarks: COCO-Stuff~\cite{caesar2018coco} and Visual Genome~(VG)~\cite{krishna2017visual}. COCO-stuff is an extension of the well known MS-COCO dataset with 80 \emph{thing} classes and 91 \emph{stuff} classes. Following \cite{sun2019image, zhao2019image, he2021context}, objects covering less than 2\% of the image are disregarded and the images with 3 to 8 objects are used here ($N_{\max}=8$). Then we have 74,777 training and 3,097 validation images of COCO-stuff. Different from COCO-stuff, Visual Genome is a dataset specifically designed for complex scene understanding and provides information of object bounding boxes, object attributes, and relationships.
Each image in VG contains 3 to 30 objects from 178 categories. Consistent with prior studies~\cite{li2021image, sun2019image}, small and infrequent objects are removed, resulting in 62,565 images for training and 5,062 for validation in the VG dataset.

\begin{table*}[ht!]
\small
\centering
\begin{tabular}{c|c|cc|cc|cc|cc} 
\toprule
\multirow{2}{*}{Resolutions}& \multirow{2}{*}{Methods}  & \multicolumn{2}{c|}{FID $\downarrow$}  & \multicolumn{2}{c|}{Inception Score $\uparrow$}&
    \multicolumn{2}{c}{Diversity Score $\uparrow$} & \multicolumn{2}{c}{CAS$\uparrow$}\\
& & COCO & VG & COCO & VG & COCO & VG & COCO & VG\\ 
\midrule
\multirow{6}{*}{$64\times 64$} & Real Images & - & - & 16.30{\scriptsize $\pm$0.40} & 13.90{\scriptsize $\pm$0.50} & - & - \\

& Layout2Im~\cite{zhao2019image} & 38.14 & 31.25 & - & - & 0.15{\scriptsize $\pm$0.06} & 0.17{\scriptsize $\pm$0.09} & 27.32 & 23.25 \\

& OC-GAN~\cite{sylvain2021object} & 29.57 & 20.27 & 10.80{\scriptsize $\pm$0.50} & 9.3{\scriptsize $\pm$0.20} & - & - & - & -\\
& Context-L2I~\cite{he2021context} & 31.32 & 33.91 & 10.27{\scriptsize $\pm$0.25} & 8.53{\scriptsize $\pm$0.13} & 0.39{\scriptsize $\pm$0.09} & 0.40{\scriptsize $\pm$0.09} & - & - \\

& LAMA~\cite{li2021image} & 19.76 & 18.11 & - & - &
0.37{\scriptsize $\pm$0.10} & 0.37{\scriptsize $\pm$0.09} & 33.23 & 30.70 \\
& LAW-Diffusion & \textbf{17.14} & \textbf{16.44} & \textbf{14.81}{\scriptsize $\pm$0.23} & \textbf{12.64}{\scriptsize $\pm$0.32} & \textbf{0.45}{\scriptsize $\pm$0.10} & \textbf{0.46}{\scriptsize $\pm$0.10} & \textbf{35.29} & \textbf{33.46} \\ 
\midrule

\multirow{6}{*}{$128\times 128$} & Real Images & - & - & 22.30{\scriptsize $\pm$0.50} & 20.50{\scriptsize $\pm$1.50} & - & - & - & - \\

& LostGAN-V2~\cite{sun2021learning} & 24.76 & 29.00 & 14.20{\scriptsize $\pm$0.40} & 10.71{\scriptsize $\pm$0.27} & 0.45{\scriptsize $\pm$0.09} & 0.42{\scriptsize $\pm$0.09} & 31.98 & 29.35 \\

& OC-GAN~\cite{sylvain2021object} & 36.31 & 28.26 & 14.60{\scriptsize $\pm$0.40} & 12.30{\scriptsize $\pm$0.40} & - & - & - & -\\
& Context-L2I~\cite{he2021context} & 22.32 & 21.78 & 15.62{\scriptsize $\pm$0.05} & 12.69{\scriptsize $\pm$0.45} & 0.55{\scriptsize $\pm$0.09} & 0.54{\scriptsize $\pm$0.09}& - & - \\

& LAMA~\cite{li2021image} & 23.85 & 23.02 & - & - &
0.46{\scriptsize $\pm$0.09} & 0.47{\scriptsize $\pm$0.09} & 34.15 & 32.81 \\

& LAW-Diffusion & \textbf{20.36} & \textbf{15.44} & \textbf{19.89}{\scriptsize $\pm$0.48} & \textbf{18.13}{\scriptsize $\pm$0.44} & \textbf{0.58}{\scriptsize $\pm$0.09} & \textbf{0.55}{\scriptsize $\pm$0.08} & \textbf{36.80} & \textbf{35.22} \\

\midrule
\multirow{5}{*}{$256\times 256$} & Real Images & - & - & 28.10{\scriptsize $\pm$1.60} & 28.60{\scriptsize $\pm$1.20} & - & - & - & -\\

& LostGAN-V2~\cite{sun2021learning} & 42.55 & 47.62 & 18.01{\scriptsize $\pm$0.50} & 14.10{\scriptsize $\pm$0.38} & 0.55{\scriptsize $\pm$0.09} & 0.53{\scriptsize $\pm$0.09} & 30.33 & 28.81 \\

& OC-GAN~\cite{sylvain2021object} & 41.65 & 40.85 & 17.80{\scriptsize $\pm$0.20} & 14.70{\scriptsize $\pm$0.20} & - & - & - & - \\

& LAMA~\cite{li2021image} & 31.12 & 31.63 & - & - &
0.48{\scriptsize $\pm$0.11} & 0.54{\scriptsize $\pm$0.09} & 30.52 & 31.75 \\

& LDM$^\dag$~\cite{rombach2022high} & 40.91 & - & - & - & - & - & - & -\\

& Frido$^\dag$~\cite{fan2022frido} & 21.67 & 17.24 & - &- &- &- & - & - \\

& TwFA~\cite{yang2022modeling} & 22.15 & 17.74 & 24.25{\scriptsize $\pm$1.04} & 25.13{\scriptsize $\pm$0.66} &
\textbf{0.67}{\scriptsize $\pm$0.00} & 0.64{\scriptsize $\pm$0.00} & - & - \\

& LAW-Diffusion & \textbf{19.02} & \textbf{15.23} & \textbf{26.41}{\scriptsize $\pm$0.96} & \textbf{27.62}{\scriptsize $\pm$0.67} & 0.63{\scriptsize $\pm$0.09} & \textbf{0.64}{\scriptsize $\pm$0.09} & \textbf{37.79} & \textbf{36.82} \\ 

\bottomrule
\end{tabular}
\vspace{2mm}
\caption{Quantitative results on COCO-stuff~\cite{caesar2018coco} and Visual Genome (VG)~\cite{krishna2017visual}. The models denoted by `$\dag$' are fine-tuned from the ones trained on a significantly larger dataset, Open-Image~\cite{kuznetsova2020open}. `-' indicates the results are not provided in their papers.}
\label{tab:comparison}
\end{table*}

\noindent\textbf{Implementation Details} \;
Following \cite{ho2020denoising, dhariwal2021diffusion}, we use $T = 1000$ and the noise magnitudes $\{\alpha_t\}_{t=1}^T$ of the diffusion process are set to linearly decrease from $\alpha_1 = 1 - 10^{-4}$ to $\alpha_T = 0.98$. Our LAW-Diffusion is trained by jointly optimizing the spatial dependency parser that generates the layout embedding $\mathcal{L}$, and the noise estimator $\tilde{\epsilon}_{\theta}(x_t, t|\mathcal{L})$ using the VLB loss defined in~\cref{eq:vlb}. We use the same diffusion training strategies and U-Net architectures as ADM~\cite{dhariwal2021diffusion}. Regarding the generation of layout embedding $\mathcal{L}$, we set the dimension of class embedding to $d_c = 32$ and the patch size of region fragments to $P = 8$. Then a two-layer MHSA with 8 attention heads is implemented as the fragment aggregation function in \cref{eq:mhsa}. Following \cite{ho2022classifier, saharia2022photorealistic}, we implement the conditional model $\epsilon_{\theta}(x_t, t|\mathcal{L})$ and unconditional model $\epsilon_{\theta}(x_t, t|\emptyset)$ in \cref{eq:classifer_free} as a single conditional model with $10\%$ probability of replacing the conditional input $\mathcal{L}$ by a learnable null embedding $\emptyset$. Due to the quadratic increase in computational overhead with the size of input images, directly generating $256\times 256$ images can be prohibitively expensive. Hence, following~\cite{esser2021taming, rombach2022high}, we utilize a VQ-VAE to downsample $256\times 256$ images to $64\times 64$, and perform our LAW-Diffusion in the compressed latent space.
For the $64\times 64$ and $128\times 128$ images, we maintain the diffusion training on image pixels. Regarding the hyper-parameters of our adaptive guidance in \cref{eq:cosine_adaptive}, we choose $\omega_{\max}=3$ and $\omega_{\min}=1$. Please refer to our supplementary materials for more implementation details.

\noindent\textbf{Evaluation Metrics} \;To comprehensively evaluate the performance of LAW-Diffusion, we adopt five metrics for quantitative comparison. Those metrics are: Inception Score~(IS)~\cite{salimans2016improved}, Fréchet Inception Distance~(FID)~\cite{heusel2017gans}, Classification Accuracy Score~(CAS)~\cite{ravuri2019classification}, Diversity Score~(DS)~\cite{zhang2018unreasonable}, YOLO Score~\cite{li2021image} and our proposed Scene Relation Score~(SRS). IS assesses the overall quality of images based on the Inception model~\cite{szegedy2016rethinking} pre-trained on ImageNet~\cite{deng2009imagenet}. FID measures the distribution distance between the synthesized images and the real ones. CAS measures the discrminative ability of generated objects and whether they can be used to train a good classifier. A ResNet~\cite{he2016deep} is trained on the objects cropped from generated images~(5 image samples are generated for each layout following~\cite{sun2021learning}) and the classification accuracy on the real objects is reported as CAS. DS reflects the diversity of generated samples. YOLO score evaluates the localization alignment between the generated objects and input bounding boxes.

\noindent\textbf{Scene Relation Score} \; Here we propose Scene Relation Score~(\textbf{SRS}) as a new metric for L2I to evaluate the rationality and plausibility of the object relations in the generated image. It is reasonable that a competent scene generator should implicitly capture the relationships among objects and the correct relations can be discovered from the synthesized images. Due to the availability of objects' bounding boxes and labels, we use the predicate classification~(PredCls) results predicted by a state-of-the-art scene graph generator to measure whether the correct relationships are captured by the image generator. Specifically, we resort to a publicly available scene graph generator, \emph{i.e.}, VCTree-EB\cite{suhail2021energy} pre-trained on Visual Genome and report the mean Recall@K(mR@K) as our Scene Relation Score~(SRS).

\subsection{Quantitative and Qualitative Comparisons}

We compare our LAW-Diffusion with the state-of-the-art L2I methods, \emph{i.e.}, Layout2Im~\cite{zhao2019image}, OC-GAN~\cite{sylvain2021object}, Context-L2I~\cite{he2021context}, LostGAN-V2~\cite{sun2021learning}, LAMA~\cite{li2021image}, TwFA~\cite{yang2022modeling}, LDM~\cite{rombach2022high} and Frido~\cite{fan2022frido}. \cref{tab:comparison} reports the quantitative comparisons for different sizes of images, in terms of FID, Inception Score~(IS), Diversity Score~(DS) and Classification Accuracy Score~(CAS). Besides, \cref{tab:yolo_srs} provides the YOLO score and the Scene Relation Score~(SRS) of different methods. For fairness, we report the performance of the compared methods from their original papers. 

With regards to the image fidelity, LAW-Diffusion significantly outperforms the existing L2I methods, achieving a new state-of-the-art performance. Especially, we observe great improvements of FID and IS scores on both COCO and VG. The noticeable improvements of the challenging CAS further verify the photo-realism of generated objects by LAW-Diffusion, so that they can be used to train a discriminative model. The comparison of SRS in \cref{tab:yolo_srs} shows that LAW-Diffusion is capable of synthesizing plausible scene images by capturing the relationships among objects.

Qualitative comparisons on COCO-Stuff and Visual Genome can be observed in \cref{fig:visual_comparison}, where the samples synthesized by different models using identical layout are presented. It is impressive that LAW-Diffusion produces perceptually appealing images with clear texture details and coherent scene relationships. Moreover, the images generated by our method faithfully complies with the spatial configurations, even in the case of large number of objects.

\begin{table}[t!]
\fontsize{6pt}{0.6\baselineskip}\selectfont
\centering
\begin{tabular}{c|c|c|c}
\toprule
\multirow{2}{*}{Resolutions} & \multirow{2}{*}{Methods} & YOLO score $\uparrow$ & Scene Relation Score~(SRS) $\uparrow$\\
& & $\text{AP}$/$\text{AP}_\text{50}$/$\text{AP}_\text{75}$ & mR@20/50/100\\

\midrule
\multirow{4}{*}{128$\times$128} & Real Images & 33.1 / 47.0 / 36.9 & 0.1652 / 0.1820 / 0.1821\\
&LostGAN-V2~\cite{sun2021learning} & 5.5 / 9.2 / 5.8 & 0.1241 / 0.1307 / 0.1295\\
&LAMA~\cite{li2021image} & 7.9 / 12.0 / 8.9 & 0.1294 / 0.1482 / 0.1489\\
& LAW-Diffusion & \textbf{14.1} / \textbf{20.6} / \textbf{17.8} & \textbf{0.1443} /  \textbf{0.1603} / \textbf{0.1631}\\

\midrule
\multirow{6}{*}{256$\times$256} & Real Images & 42.9  / 60.2 / 48.2 & 0.1703 / 0.1927 / 0.1932\\
&LostGAN-V2~\cite{sun2021learning} & 9.1 / 15.3 / 9.8 & 0.1241 / 0.1307 / 0.1295\\
&LAMA~\cite{li2021image} & 13.4 / 19.7 / 14.9 & 0.1260 / 0.1321 / 0.1333\\
&Frido~\cite{fan2022frido} & - / 30.4 / - & 0.1375 / 0.1535 / 0.1578\\
&TwFA~\cite{yang2022modeling} & - / 28.2 / 20.1 & 0.1407 / 0.1474 / 0.1487\\
& LAW-Diffusion & \textbf{21.5} / \textbf{34.2} / \textbf{23.4} & \textbf{0.1485} / \textbf{0.1742} / \textbf{0.1750}\\
\bottomrule
\end{tabular}
\vspace{2mm}
\caption{Comparisons of YOLO score and SRS.}
\label{tab:yolo_srs}
\end{table}

\begin{table}[t!]
\scriptsize
\centering
\begin{tabular}{c|c|ccc} 
\toprule
\multirow{2}{*}{Methods} & \multirow{2}{*}{IS $\uparrow$} & \multicolumn{3}{c}{Scene Relation Socre (SRS) $\uparrow$} \\

&  & mR@20 & mR@50 & mR@100 \\
\midrule

$\omega = 1$ & 13.93{\tiny $\pm$0.31} & 0.1271 & 0.1233 & 0.1316 \\
$\omega = 3$ & 16.62{\tiny $\pm$0.49} & 0.1419 & 0.1484 & 0.1488 \\
$\omega = 5$ & 15.95{\tiny $\pm$0.31} & 0.1324 & 0.1401 & 0.1438 \\
$\omega_t: 1\tiny{\nearrow}3$ & 15.21{\tiny $\pm$0.37} & 0.1319 & 0.1345 & 0.1370\\
$\omega_t: 1\tiny{\nearrow}5$ & 14.68{\tiny $\pm$0.28} & 0.1302 & 0.1310 & 0.1334 \\
$\omega_t: 3\tiny{\searrow}1$ & 18.13{\tiny $\pm$0.44} & \textbf{0.1443} & \textbf{0.1603} & \textbf{0.1631}\\
$\omega_t: 5\tiny{\searrow}1$ & \textbf{18.24}{\tiny $\pm$0.29} & 0.1392 & 0.1436 & 0.1544 \\
\midrule
$\omega = 1$(w/o loc) & 9.69{\tiny $\pm$0.32} & 0.1168 & 0.1206 & 0.1257 \\
$\omega = 3$(w/o loc) & 12.34{\tiny $\pm$0.58} & 0.1235 & 0.1287 & 0.1298 \\
$\omega = 5$(w/o loc) & 13.97{\tiny $\pm$0.32} & 0.1214 & 0.1241 & 0.1260 \\
$\omega_t: 1\tiny{\nearrow}3$(w/o loc) & 12.06{\tiny $\pm$0.42} & 0.1198 & 0.1264 & 0.1278\\

$\omega_t: 1\tiny{\nearrow}5$(w/o loc) & 11.42{\tiny $\pm$0.41} & 0.1190 & 0.1221 & 0.1259 \\
$\omega_t: 3\tiny{\searrow}1$(w/o loc) & 14.78{\tiny $\pm$0.33} & 0.1252 & 0.1315 & 0.1327\\

$\omega_t: 5\tiny{\searrow}1$(w/o loc) & 14.52{\tiny $\pm$0.25} & 0.1263 & 0.1334 & 0.1358 \\

\bottomrule
\end{tabular}
\vspace{2mm}
\caption{Ablation study on VG 128$\times$128.}
\label{tab:ablation}
\end{table}

\subsection{Instance-level 
Reconfiguration}
\begin{figure}[t!]
\begin{center}
\includegraphics[width=\linewidth]{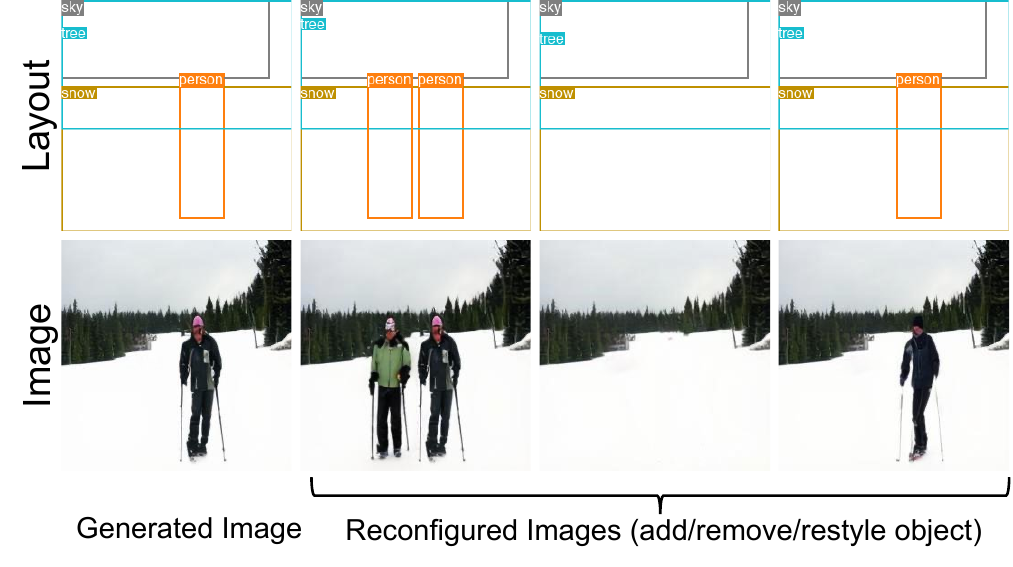}
\end{center}
\caption{An example of instance-level reconfiguration by LAW-Diffusion. Three types of reconfiguration are shown in this figure (adding/removing/restyling a person in the generated image). Plausible results are obtained using layout-aware latent grafting.}
\vspace{-5.5mm}
\label{fig:result_graft}
\end{figure}

As presented in \cref{sec:reconfigure}, a trained LAW-Diffusion has flexible instance-level controllability, involving the abilities of adding/removing/restyling an instance in the generated scene while preserving the other contents. An example of these three types of reconfiguration is given in \cref{fig:result_graft}. The reconfigured images look plausible and well preserve the coherence in the scene, thus verifying the effectiveness of our proposed layout-aware latent grafting mechanism.

\subsection{Ablation Study}

To verify the effectiveness of proposed location-aware cross-object attention and adaptive guidance schedule, we conduct ablation experiments on VG 128$\times$128 in \cref{tab:ablation}. Here, we first introduce a baseline variant of LAW-Diffusion, dubbed LAW-Diffusion (w/o loc), which replaces the location-aware attention by a location-agnostic but class-aware attention used in prior works~\cite{he2021context, yang2022modeling}. Specifically, we use the MHSA layers similar to what we used in \cref{sec:LAW} to augment each object's class embedding $c_i$ with only contextual class-aware information. Then the transformed object representations are filled into their bounding boxes and aggregated as the layout embedding using average pooling. In this way, it only captures class-aware relationships and is also used to guide the generation of diffusion model. Moreover, \cref{tab:ablation} shows the results of LAW-Diffusion and LAW-Diffusion (w/o loc) with different guidance strategies. For example, $\omega_t: 3\tiny{\searrow}1$ means LAW-Diffusion with cosine annealed guidance scale from $\omega_{\max}=3$ to $ \omega_{\min}=1$, and $\omega = 3$(w/o loc) denotes LAW-Diffusion (w/o loc) using a fixed guidance scale $\omega=3$. Similarly, $\omega_t: 1\tiny{\nearrow}3$ and $\omega_t: 1\tiny{\nearrow}5$ denotes the increasing guidance scales.

By comparing IS and SRS between the variants of LAW-Diffusion and LAW-Diffusion (w/o loc) in \cref{tab:ablation}, we can conclude that our location-aware cross-object attention can both improve the generated fidelity and capture the reasonable relations among objects. Besides, it is clear that our adaptive guidance schedule promotes the improvement of the IS scores of generated images. Considering both image fidelity and rationality of the object relations, we select LAW-Diffusion with cosine annealing guidance $\omega_t:3\tiny{\searrow}1$ as our final model. Please refer to our supplementary materials for more ablation studies and human evaluations.

\section{Conclusion}
In this paper, we present a semantically controllable Layout-AWare diffusion model, termed LAW-Diffusion to generate complex scenes from compositional layout configurations. Specifically, we propose a location-aware cross-object attention module to learn a layout embedding encoding the spatial dependencies among objects. Further, an adaptive guidance schedule is introduced for the layout guidance to maintain both layout semantic alignment and object's texture fidelity. Moreover, we propose a layout-aware latent-grafting mechanism for instance reconfiguration on the generated scene. Furthermore, a new evaluation metric for L2I, dubbed Scene Relation Score~(SRS) is proposed to measure how the images preserves rational relations. Extensive experiments show that our method yields the state-of-the-art generative performance, especially with coherent object relations. 

\noindent\textbf{Limitation and future work} With regards to the limitation of our work, we only focus on the task of Layout-to-Image generation whose object categories are pre-defined,fixed, and closed-world. Additionally, current version fails to specify scene-level style and semantics with global scene description. In future, we aim to combine our LAW-Diffusion with RegionCLIP~\cite{zhong2022regionclip} to achieve open-vocabulary L2I generation, where the objects generated in the scene can belong to arbitrary novel categories and both object-level and scene-level fine-grained semantic controls can be achieved.

\section{Acknowledgement}
This work was supported in part by National Key R$\&$D Program of China under Grant No.2021ZD0111601, National Natural Science Foundation of China (NSFC) under Grant No.61836012, U1811463, U21A20470, 62006255, 61876224, 62206314, GuangDong Basic and Applied Basic Research Foundation under Grant No.2017A030312006, 2022A1515011835, 2023A1515011374 (GuangDong Province Key Laboratory of Information Security Technology)

\clearpage
{\small
\bibliographystyle{ieee_fullname}
\bibliography{egbib}
}

\end{document}